\documentclass{article}

\usepackage{microtype}
\usepackage{graphicx}
\usepackage{subcaption}
\usepackage{booktabs}

\usepackage{hyperref}

\usepackage[accepted]{icml2026}

\usepackage{amsmath}
\usepackage{amssymb}
\usepackage{mathtools}
\usepackage{amsthm}

\usepackage[capitalize,noabbrev]{cleveref}

\theoremstyle{plain}

\theoremstyle{definition}

\theoremstyle{remark}

\usepackage[textsize=tiny]{todonotes}

\begin{document}

\twocolumn[
  \icmltitle{Position: Reinforcement Learning Foundation Models \\
  Should Already Be A Thing}
  \icmltitlerunning{Position: Reinforcement Learning Foundation Models Should Already Be A Thing}

  \icmlsetsymbol{equal}{*}

  % \todo[inline]{Replace placeholder author block before camera-ready.}
  \begin{icmlauthorlist}
    \icmlauthor{Abdelrahman Zighem}{ENS,SODA}
    \icmlauthor{Jill-Jênn Vie}{SODA}
  \end{icmlauthorlist}

  \icmlaffiliation{ENS}{École normale supérieure de Paris, PSL University, Paris, France}
  \icmlaffiliation{SODA}{Soda team, Inria Saclay, Palaiseau, France}

  \icmlcorrespondingauthor{Abdelrahman Zighem}{abdelrahman.zighem@ens.psl.eu}

  \icmlkeywords{Foundation Models, Reinforcement Learning,
    Tabular Foundation Models, Graph Foundation Models,
    Prior-Fitted Networks, In-Context Learning}

  \vskip 0.3in
]

\printAffiliationsAndNotice{}

\begin{abstract}

Foundation models for language and vision are powered by internet-scale data,
while structured domains such as tabular prediction are powered by synthetic data.
This substitute shifts the challenge from collection to prior design. Such 
priors already exist for many structured tasks: TabPFN and its successors solve 
tabular classification with a transformer pretrained on a synthetic Bayesian prior.

We make two points. \textbf{First}, reinforcement learning is the conspicuous gap: 
sampling a synthetic MDP is as feasible as sampling a synthetic tabular dataset,
yet no in-context RL work treats prior design as a primary objective. 
\textbf{Second}, MDPs admit a fixed-size sufficient statistic, independent of 
the episodes observed and tabular in shape, which makes them directly amenable 
to the attention-based architectures used for tabular foundation models, with a
policy head replacing the supervised target. Together these define the agenda 
for an RL foundation model.

As a proof of concept, we train a Graph Attention Network entirely on synthetic 
MDPs and show that, with no task-specific tuning, it solves held-out tabular 
benchmarks in context, both online and offline: online, in far fewer episodes 
than UCB-VI and tabular Q-learning, and offline, competitively with VI-LCB.
\end{abstract}

\section{Foundation Models Need Strong Priors}
\label{sec:intro}

The foundation-model paradigm (pretrain on a vast distribution, adapt to
specific instances) has spread well beyond
language~\citep{bommasani2021foundation}. TabPFN, TabICL and their
successors~\citep{hollmann2023tabpfn,hollmann2025tabpfnv2,qu2025tabicl,
qu2026tabiclv2} showed that tabular classification problems can be solved by 
transformers trained on a synthetic prior over Bayesian classifiers. 
Prior-fitted network (PFN) training~\citep{mueller2022pfn} has since been 
argued to be the dominant paradigm for Bayesian inference in any domain where a 
tractable prior can be sampled~\citep{mueller2025position}.

The recipe has three requirements: a data distribution that is cheap to sample,
expressive enough to cover the deployment regime, and sufficiently structured
for learning useful inductive biases. 
No such prior exists for reinforcement learning, so building
one is an important next step for the field.

\section{The Missing Prior for Reinforcement Learning}
\label{sec:gap}

There is already solid theoretical support for the idea that in-context 
reinforcement learning is a viable and powerful method, even if its practical 
applications are still being explored or developed. For instance, 
\citet{lin2024transformers} prove that transformers with ReLU attention can
efficiently implement near-optimal online RL algorithms (UCB-VI, Thompson
sampling, LinUCB) purely in context, via supervised pretraining on offline
trajectories. What the field has not provided is a prior that makes pretraining
useful across arbitrary MDPs.

The question of what prior to place over MDPs is not new.
\citet{strens2000bayesian} and \citet{dearden1998bayesian} put it at the
center of Bayesian RL, proposing Gaussian reward priors and Dirichlet
transition priors as tractable choices. That thread has been almost entirely
dropped by the modern in-context RL and meta-RL
literature~\citep{wang2016learning,laskin2023algorithm,schiff2025tabpfnrl,
lee2023supervised,grigsby2024amago,son2025dicp,lin2024transformers}.
One partial exception is \citet{duan2016rl2}, whose experimental setup
does sample random MDPs from a Gaussian/Dirichlet prior, but this point is 
only briefly discussed in the paper.

Some recent work uses LLMs as priors~\citep{choi2022lmpriors,yan2025llmrl},
querying them to propose reward functions or action distributions. This shifts
rather than solves the problem: a full LLM forward pass per sample makes
large-scale synthetic pretraining impractical.

All these papers use trajectories as training data, from hand-picked
task families (bandit suites, procedurally generated platform 
games~\citep{cobbe2020procgen}, robotic-control benchmarks), producing models 
adapted to a specific environment family rather than a foundation model in the 
sense of~\citet{bommasani2021foundation}. Sequence-modeling approaches such as 
the  Decision Transformer~\citep{chen2021decision} and Gato~\citep{reed2022gato} 
likewise treat data as given. None of these works ask what distribution over 
environments would make pretraining optimal.

There is a second, independent reason to move away from raw trajectories: they
do not scale. Recurrent approaches are bounded by vanishing and exploding
gradients over long sequences~\citep{bengio1994learning}, which caps the
effective memory regardless of model size. Transformer approaches avoid that
failure mode but introduce another: attention cost is quadratic in context
length~\citep{vaswani2017attention}, and the context for $N$ episodes of
horizon $H$ has $\Theta(NH)$ tokens. As experience accumulates, managing this 
growing context becomes increasingly
difficult, and the part of the history most at risk is the early, exploratory
phase, which carries the most information about the MDP. Context compression
rather than context truncation is an active research direction. Recent work
on end-to-end compression for tabular foundation models~\citep{zabergja2026taco}
shows that latent compression can reduce memory and inference costs by orders of
magnitude, so the problem may eventually become more tractable. For
now it remains a genuine obstacle, especially in the case of RL where
trajectories can be numerous, long, and of various length.

\section{MDPs as Graphs, and Their Tabularization}
\label{sec:tab}

A Markov Decision Process (MDP) $M = (\mathcal{S}, \mathcal{A}, P, r, \gamma)$
is a graph: states are nodes, actions label outgoing edges, and the Markov 
kernel $P(\cdot \mid s, a)$ gives weighted adjacency. Existing in-context RL
approaches feed the model the \emph{trajectory} traced over this graph, a
sequence of length $\Theta(NH)$ for $N$ episodes of average horizon $H$.

For small finite MDPs, the Markov property completely eliminates the
need for trajectory-length context. The trajectories with a tabular MDP collapse
into three sufficient statistics: visit counts $N(s,a) \in \mathbb{N}$ for each 
pair, empirical mean rewards $\hat{r}(s,a) \in [0,1]$ and empirical transition 
row $\hat{P}(\cdot \mid s, a) \in \Delta(\mathcal{S})$, with the convention that
unvisited rows default to uniform transitions and rewards are initialized at 
$0$. Stacked across $(s,a)$ pairs, these form a matrix
\begin{equation}
    Z \;=\; \bigl(\, N(s,a),\;\hat{r}(s,a),\;
    \hat{P}(\cdot \mid s,a) \,\bigr)_{(s,a) \in \mathcal{S} \times \mathcal{A}}
\end{equation}
of shape $|\mathcal{S}||\mathcal{A}| \times (|\mathcal{S}| + 2)$, with total
size $\mathcal{O}(|\mathcal{S}|^2 |\mathcal{A}|)$ and crucially
\emph{independent of $N$ and $H$}. Each row of $Z$ is one state--action pair.
The visit count and empirical reward are scalar features. The empirical
transition row is an $|\mathcal{S}|$-dimensional feature vector that encodes
graph adjacency and edge weights. This is the format consumed by a
permutation-equivariant set-transformer of the TabPFN family.

When the state space grows, or when it is infinite, the cost of storing 
$\mathcal{O}(S^2 A)$ entries becomes unmanageable. One can imagine various 
options to generalize this approach: mesh discretization and clustering are two
obvious candidates. However, discretization suffers from the curse of
dimensionality, and clustering may require subtle feature engineering when the 
data is hard to separate. We leave further exploration in this direction to 
future work.

Adopting the tabular \emph{representation} does not mean adopting TabPFN as a 
model. PFN-style in-context learning is supervised: it predicts a target column 
of the input given the others. The output of the RL model is a policy over 
actions. The tabular framing provides the input format, the synthetic-prior 
training recipe, and some architectural ideas, but not the loss or the output 
heads. We turn to those next.

\section{A Foundation Model for small tabular MDPs}
\label{sec:arch}
In this section, we describe an architecture for solving small RL problems 
that builds upon the ideas described 
in this paper. The implementation, training code, and evaluation scripts 
are available at
\url{https://github.com/Shika-B/One-Shot-Reinforcement-Learning}.

\subsection{Design}
A Graph Attention Network (GAT, \citet{veličković2018graphattentionnetworks}) 
takes $[\log(1+N), \hat{r}]$ as features and $\hat{P}$ as
multiplicative attention-biases, produces a per-row representation of the
features and exposes a policy head that outputs, for a given state $s$, an
estimate $\pi_{\mathrm{model}}(s)$ of the probability vector $\pi^\star(s)$. The
representation is refined by $K$ propagation layers, each a single round of
message passing along the transition structure. These layers are weight-tied, so
$K$ is not baked into the parameters: it can be increased or decreased at
evaluation time to spend more or less compute on planning depth, without
retraining. Permutation equivariance over rows gives automatic invariance under
relabelings of states and actions. No positional encoding is used, so the
integer identity of a given state in one MDP does not leak to another. Since MDPs
in a batch differ in size, states and actions are padded to fixed maxima
$S_{\max}$ and $A_{\max}$, and a state/action mask keeps this padding out of every
attention, update, and policy output (cf. Appendix~\ref{app:impl}).

Using a GAT allows the model to be structurally aware of the underlying graph
structure of the MDP, rather than learn it. Each round of message 
passing propagates information about state-action pairs backward through the 
graph: when the model learns that a pair $(s, a)$ is a good
target, it should also raise its estimate of the states from which $(s, a)$ is
easily reachable.

Given a sampled MDP $M$, \citet{mueller2022pfn} argues that the model should 
estimate $\mathbb{E}_{M' \sim \mathbb{P}(M' \mid Z)}
\left[\pi_{M'}^* ( \cdot )  \mid Z\right]$ to optimize for Bayesian inference.

We choose to directly regress against the optimal policy of the \emph{sampled}
MDP $M$. Concretely the target is its Boltzmann-optimal policy
$\pi^*_{M,\tau} = \mathrm{softmax}(Q_M^* / \tau)$, a softened greedy policy that
recovers the deterministic $\pi_M^*$ as $\tau \to 0$ (cf. Appendix~\ref{app:impl}).
This is an unbiased estimate of the above quantity: since $Z$ is generated by
exploring $M$,  the true $M$ is already a posterior sample
$M\sim P(M\mid Z)$, so regressing onto $\pi^*_{M,\tau}$ given $Z$ is the same
generative process as regressing onto $\pi^*_{M',\tau}$ for a fresh
$M'\sim P(M\mid Z)$. Both yield the Bayes-optimal
$\mathbb{E}[\pi^*_\tau ( \cdot )  \mid Z]$.
Note that this is the optimal target in an \emph{offline setting}, and the 
optimal exploration strategy may differ significantly. We observe empirically 
that such a target still delivers solid performance when used in an online 
setting.

Each training batch is a collection of freshly sampled MDPs. We draw every MDP
from a broad prior that varies its size (from two to a few dozen states, with two
to four actions), its connectivity, its transition stochasticity, and its reward
structure. Connectivity is kept sparse: each state has a small outdegree, and its
successors are usually the nearest neighbors in a latent geometry (a chain, a
grid, a mesh, or an unstructured random graph) so that the transition graph
has a coherent layout rather than arbitrary edges. Each 
transition row is drawn from a Dirichlet whose concentration spans 
near-deterministic to fully diffuse dynamics,
so a single prior covers both clean and noisy environments. Rewards are sparse
and assigned per state-action pair, with magnitudes and signs drawn
independently, so per-step penalties and a sparse positive reward can coexist in
the same MDP. Sparsity is applied explicitly: each $(s,a)$ reward is independently
zeroed with probability $1 - p_{\text{keep}}$, where $p_{\text{keep}}$ is itself
resampled for every MDP, so the fraction of nonzero rewards varies across the 
prior. We solve each sampled MDP exactly with value iteration to obtain its
optimal policy, then simulate a finite budget of exploration to produce the noisy
statistics $Z$ the model actually sees, and train the network to recover the
optimal policy from $Z$ alone. The full distributions and constants are given in
Appendix~\ref{app:impl}.

\subsection{Results}

\begin{figure}[ht]
  \begin{center}
    \centerline{\includegraphics[width=\columnwidth]{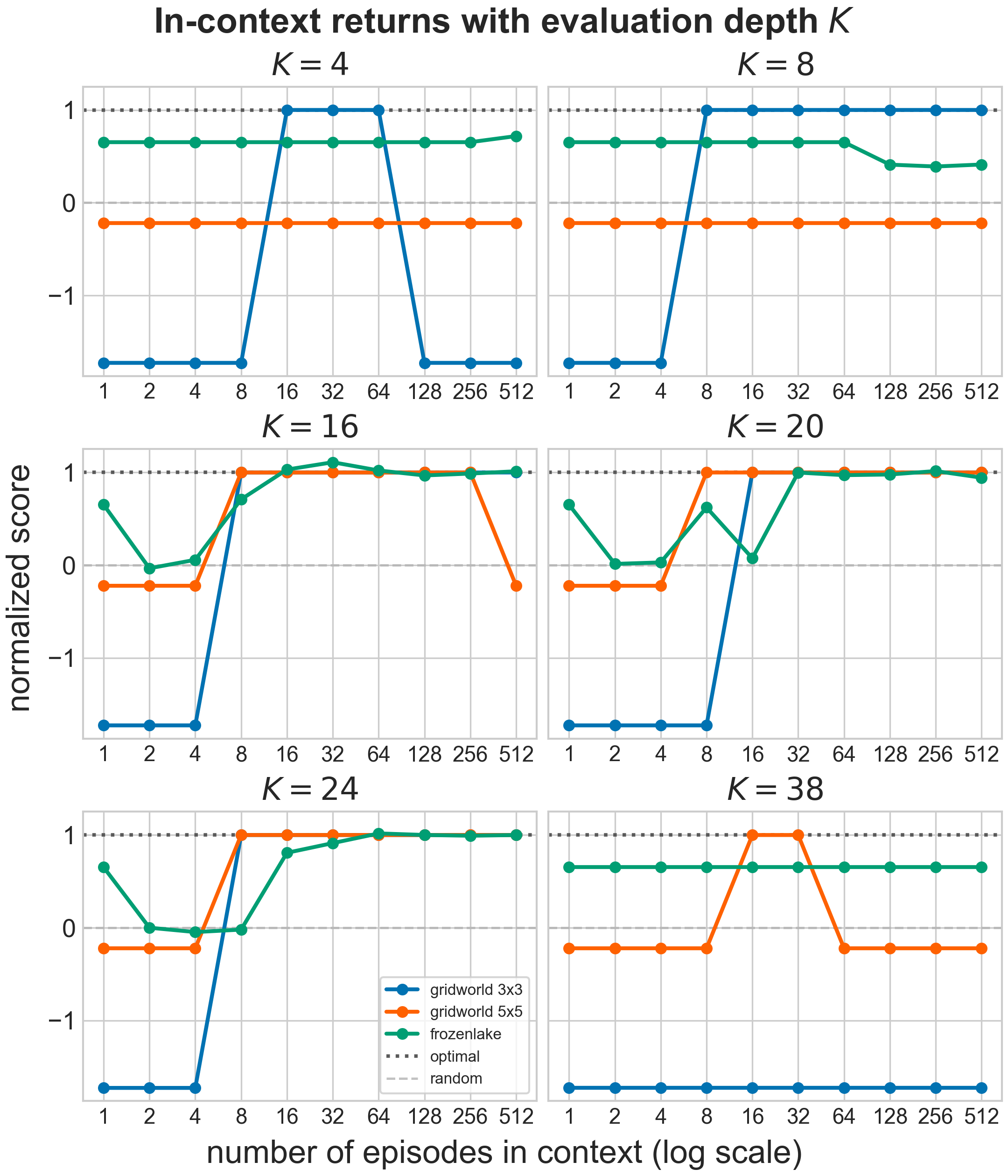}}
    \caption{
      In-context returns as a function of the number of episodes in context, one
      panel per evaluation depth $K$ and one line per held-out environment. The
      model was trained with $K = 20$. Since FrozenLake is a stochastic
      environment, the return is an average over 256 runs. Normalized scores
      are computed as
      $\frac{R - R_{\mathrm{rand}}}{R_{\mathrm{opt}} - R_{\mathrm{rand}}}$
      where $R$ is the average model return, $R_{\mathrm{opt}}$ is the optimal
      average return and $R_{\mathrm{rand}}$ is the average return under a
      random policy.
    }
    \label{K_evals}
  \end{center}
\end{figure}

\begin{figure}[ht]
  \begin{center}
    \centerline{\includegraphics[width=\columnwidth]{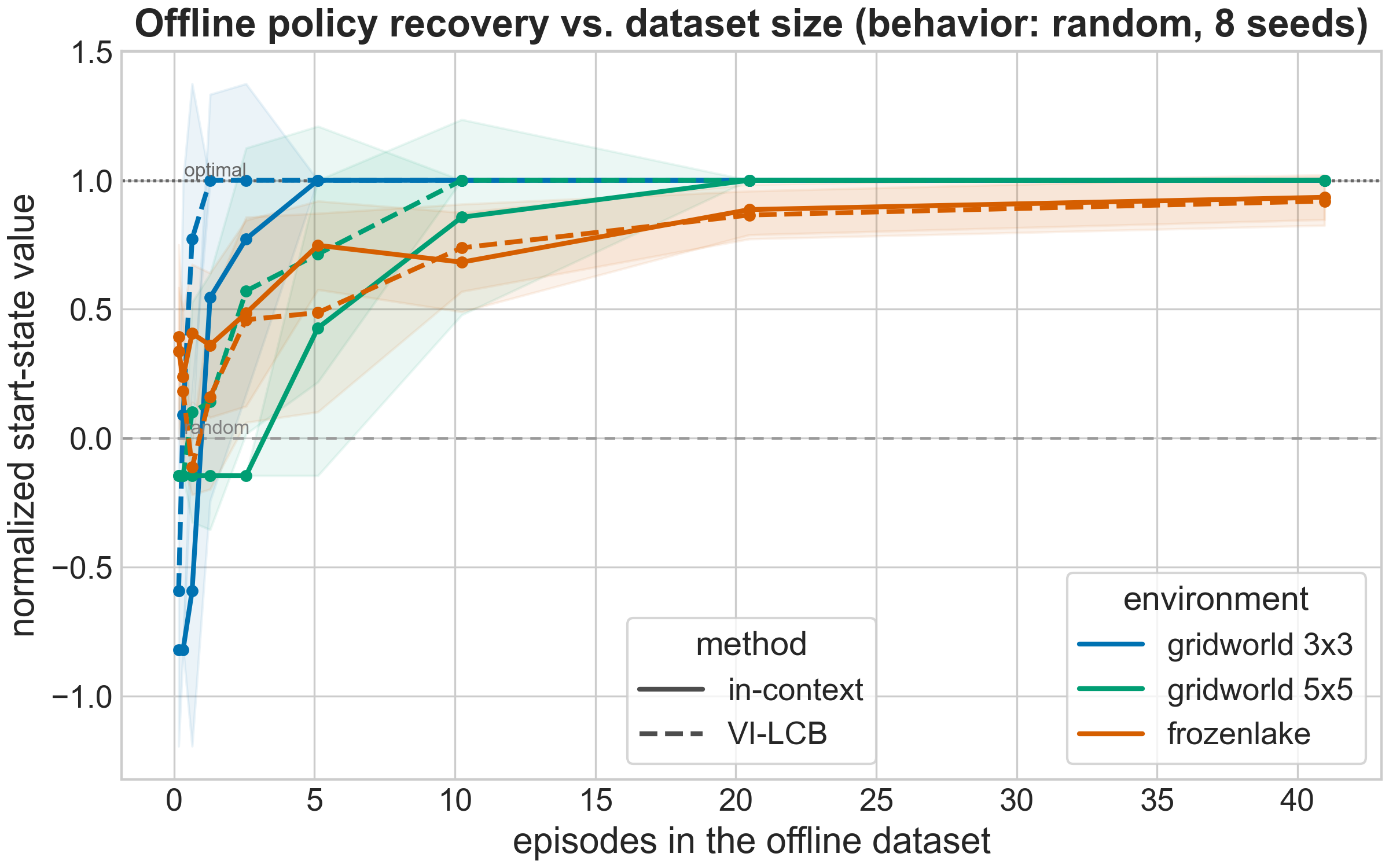}}
    \caption{Offline policy recovery from a fixed uniform-random dataset. The
    figure plots the exact normalized start-state value of the policy recovered
    by the in-context model (solid) and by pessimistic value iteration
    (VI-LCB, dashed) against the number of episodes in the dataset, for three
    held-out MDPs (color); mean over $8$ seeds (shaded: one standard
    deviation). The VI-LCB penalty coefficient is $c = 0.1$, hand-picked to
    maximize FrozenLake performance.}
    \label{fig:offline}
  \end{center}
\end{figure}

\begin{table}[ht]
\centering
\begin{tabular}{lccc}
\hline
\quad & In-Context & UCB-VI & Q-Learning \\
\hline
GridWorld 3x3 & 6  & 16 & 469  \\
GridWorld 5x5 & 6  & 53 & 1205 \\
FrozenLake    & 27 & 46 & 1014 \\
\hline
\end{tabular}
\caption{Median number of episodes it takes to converge, on 12 seeds.}
\label{tab:qlearning_ucb}
\end{table}

Although the model is trained with a purely offline objective, it can also be 
used to guide its own exploration autoregressively. This is the evaluation 
protocol we used in Figure \ref{K_evals} and Table \ref{tab:qlearning_ucb}.
The full description of the evaluations is
given in Appendix~\ref{app:eval}. 

The two benchmarks used here are
GridWorld~\citep{sutton2018reinforcement} and 
FrozenLake~\citep{brockman2016openai}. GridWorld's deterministic 
environment is a $k \times k$ grid where each step taken has a negative rewards 
until the agent finds the self-absorbing goal case, which has a strong positive reward.
FrozenLake is a highly stochastic analogue to GridWorld, with holes scattered 
across the grid and each action having a chance of making you ``slip`` to 
neighboring cases (which may be holes). Both holes and the goal are 
self-absorbing states, with strong positive and negative rewards respectively. 

Our architecture allows for tunable depth at test time and we take advantage of 
that. Figure \ref{K_evals} shows that the model's performance degrades 
significantly as $K$ diverges from $20$, which is the value it was trained on. 
However, slight increases lead to improvement over the initial value, as is
the case for $K = 24$ here. Consequently, we use $K = 24$ for the other
evaluations.

Figure \ref{fig:offline} isolates the model as a purely offline estimator.
We compare our model to the VI-LCB offline RL algorithm \citep{rashidinejad2021bridging}, 
which performs pessimistic value iteration using lower-confidence-bound estimates.
The fixed dataset is gathered by a uniform-random policy, both the model and 
VI-LCB must recover a policy from the same statistics, scored by the exact 
value at the start state. Both reach the optimum as the dataset grows, 
and on the deterministic GridWorlds a handful of episodes already suffice. 
The comparison is deliberately generous to the baseline, since the VI-LCB
penalty was tuned on FrozenLake, yet the in-context model stays competitive 
throughout and recovers strong policies from very little data. FrozenLake is the
hard case for both methods: under uniform exploration the goal sits behind the 
holes and is observed only rarely, so the dataset is coverage-limited and 
convergence is slower.

Table \ref{tab:qlearning_ucb} shows that the model solves these problems with
far fewer episodes than standard RL algorithms such as Q-learning
\citep{watkins1992qlearning}, and even than a model-based algorithm such as
UCB-VI \citep{azar2017minimax}, which is the upper-confidence-bound analogue of 
VI-LCB, often used for online RL.

Overall, it is pretty remarkable that the model generalizes to new unseen 
scenarios.

\section{Limitations, Open Problems and Future Work}
\label{sec:limits}

Our prototype is deliberately minimal, and several of its assumptions are also
its limitations. We group them by the direction in which we expect them to be
relaxed.

\paragraph{State features.} The model is blind to the
identity of a state by construction: the rows of $Z$ are permutation-equivariant
and carry no per-state features, so a state's integer index means nothing and
cannot leak between MDPs. That buys invariance to
relabeling at the cost of geometry. Two physically adjacent states look unrelated
to the model until it observes a transition between them, and a state it has
never visited is a uniform-prior blank with no neighbors to borrow from. 
One extension is to replace the bare index with a feature vector
$x_s \in \mathbb{R}^d$ per state (and per action), which gives the input a notion
of distance between states. The model could reach an
unseen state by extrapolating from nearby, better-observed ones instead of
leaning on the prior alone, the same way a tabular foundation model generalizes 
across feature space rather than row indices. This asks the prior to generate
informative features whose distances track similarity in dynamics and value. Our
generative process already places states in a latent geometry to wire up
transitions (cf. Appendix~\ref{app:impl}), and feeding that geometry to the 
model, not just the edges it induces, is a solid place to start.

\paragraph{Continuous states and actions.} Enumerable $(s,a)$ pairs are the core
assumption of the tabular reduction, and large discrete or continuous state
spaces break it. Feature-indexed rows point the way out on the state side: a
dataset of observed $(s, a, s', r)$ tuples keyed by features rather than indices
is exactly the tabular shape our model already consumes. The action side
needs a different head. Our masked softmax over a finite action set does not
extend to continuous control; the head can instead emit the
parameters of a squashed Gaussian, a Gaussian passed through a $\tanh$ to respect
bounded action ranges, which is the standard continuous-action policy
parameterization popularized by soft actor-critic~\citep{haarnoja2018sac}. This
keeps our design choice of predicting a policy directly rather than action values.

\paragraph{Size generalization.} Even with features, transfer across scales stays
hard. A model trained on $|\mathcal{S}| \approx 50$ and deployed at
$|\mathcal{S}| \approx 500$ meets graphs with different degree distributions,
diameters and spectral gaps, so the dynamics that govern behaviors completely
shift rather than merely grow. This is not a capacity problem; it likely needs an
explicit scale prior in the data, or a representation invariant to size in a
stronger sense than permutation equivariance.

\paragraph{Amortizing the online cost.} Each environment step updates one row of
$Z$ and one count in $\mathcal{O}(1)$, already far cheaper than the $\Theta(NH)$
re-encoding of trajectory-based models. We currently re-encode all of $Z$ to
replan once per episode; whether the per-step update can instead be made
incremental, refreshing only the affected rows of the propagation, is open and
would make fully online deployment cheap.

\paragraph{Beyond the Markov statistic.} The whole approach rests on the Markov
property: $Z$ is a sufficient statistic only when the environment is Markov in
the observed state. Partial observability breaks this, and recovering a latent
state would reintroduce the sequence-modeling costs we set out to avoid.
Relatedly, the offline study (Figure~\ref{fig:offline}) shows the method is only
as good as the dataset's coverage: under poor exploration the prior carries the
estimate, so a key open question is when the prior helps and when it misleads,
i.e. how robust the method is to prior misspecification.

\section{Conclusion}

We have argued that prior design deserves to be a primary objective in
in-context RL, and that the Markov property already provides the right
representation for small problems: a fixed-size sufficient-statistics matrix 
that gives RL data a tabular shape amenable to non-sequential architectures. 
Open problems remain, but the broader foundation-model paradigm has taken hold 
in one domain after another by treating prior design as a first-class concern, 
and there are solid reasons to expect the same here.

\section*{Use of Generative AI}
Generative AI was used to help with formatting and grammar checks.

\section*{Impact Statement}

This paper presents work whose goal is to advance the field of Machine
Learning. There are many potential societal consequences of our work, none
of which we feel must be specifically highlighted here.

\bibliography{references}
\bibliographystyle{icml2026}

\newpage
\appendix
\onecolumn
\section{Implementation and training details}
\label{app:impl}

This appendix documents the proof-of-concept implementation: the prior over
MDPs, the supervision targets, the model input, the architecture, and the
optimization. The four stages map one-to-one onto the source files
\texttt{prior.py}, \texttt{train.py}, \texttt{model.py} and
\texttt{evaluation.py}.

\subsection{Prior over MDPs}

We sample finite MDPs $M = (\mathcal{S}, \mathcal{A}, P, r, \gamma)$ with
controlled variation in size, connectivity, transition stochasticity and reward
structure; the discount is fixed at $\gamma = 0.95$. State and action counts are
drawn per MDP,
\begin{align*}
  &S \sim \mathrm{LogUniform}(2, 32) \\
  &A \sim \mathrm{Uniform}(2, 4) \\
  &O \sim \mathrm{PowerLaw}(o) \propto 1/o
\end{align*}

with an outdegree $O \in [\![1, 6]\!]$ shared across the states of a given MDP.
Each MDP draws a latent geometry $g \in \{\text{chain}, \text{grid},
\text{mesh}, \text{random}\}$; states are placed as points $x_s \in [0,1]^d$
($d = 1, 2, 3$ for the geometric cases), and the successor set
$\mathcal{N}(s,a)$ of size $O A_{\max}/2$ is taken as the nearest neighbors in
this latent space, or uniformly at random when there is no geometry.

Transition rows are Dirichlet over a size-$O$ support drawn from
$\mathcal{N}(s,a)$,
\begin{align*}
  &P(\cdot \mid s,a) \sim \mathrm{Dirichlet}(\alpha \mathbf{1}_O) \\
  &\alpha \sim \mathrm{LogUniform}(0.05, 5.0),
\end{align*}
so the concentration $\alpha$ moves the dynamics between near-deterministic
($\alpha \ll 1$) and diffuse ($\alpha \gg 1$). Rewards are fixed per $(s,a)$,
sparsified by a per-MDP keep probability and signed independently per pair:
\begin{align*}
  &r(s,a) = \mathbf{1}\{u_{s,a} < p_{\text{keep}}\}\,\sigma_{s,a}\,z_{s,a} \\
  &z_{s,a} \sim \mathrm{Beta}(2, 5) \\
  &\Pr(\sigma_{s,a} = +1) = p_{\text{pos}}
\end{align*}

with $p_{\text{keep}} \sim \mathrm{Beta}(2, 4)$ and the positive fraction
$p_{\text{pos}} \sim \mathrm{Uniform}(0,1)$ drawn once per MDP. Signing each pair
separately, rather than fixing one sign per MDP, lets the prior cover mixed-sign
reward structures, where per-step penalties and a sparse reward coexist in the
same environment. The constants are
listed in Table~\ref{tab:prior}.

\begin{table}[h]
  \caption{Prior and model hyperparameters.}
  \label{tab:prior}
  \centering
  \small
  \begin{tabular}{lll}
    \toprule
    Symbol & Meaning & Distribution / value \\
    \midrule
    $\gamma$ & discount & $0.95$ \\
    $S$ & states & $\mathrm{LogUniform}(2, 32)$ \\
    $A$ & actions & $\mathrm{Uniform}(2, 4)$ \\
    $O$ & outdegree & $\mathrm{PowerLaw},\ o \in [\![1,6]\!]$ \\
    $\alpha$ & concentration & $\mathrm{LogUniform}(0.05, 5.0)$ \\
    $g$ & latent geometry & uniform on geometries \\
    $p_{\text{keep}}$ & reward keep probability & $\mathrm{Beta}(2, 4)$ \\
    $z_{s,a}$ & reward magnitude & $\mathrm{Beta}(2, 5)$ \\
    $p_{\text{pos}}$ & positive-reward fraction & $\mathrm{Uniform}(0,1)$ \\
    $\beta$ & attention regularization & $1.0$ \\
    \bottomrule
  \end{tabular}
\end{table}

\subsection{Supervision targets}

For each sampled MDP we run value iteration on the true dynamics ($500$ sweeps)
to obtain the optimal action value $Q^*(s,a)$, from which the supervision target
is built. Both $Q^*$ and the observed rewards are divided by the per-MDP reward
scale $\max_{s,a} |r(s,a)|$, so the target is reward-scale invariant and the
eval-time normalization (a running maximum) lines up with what the model saw in
training.

\subsection{Model input}

The model never sees the true dynamics. At each step we simulate a finite
exploration budget against $(P, r)$: a per-MDP mean count is drawn log-uniformly
and a Poisson number of visits is then sampled for each $(s,a)$, so coverage
spans the near-zero-data regime, where empirical rows fall back to the prior, up
to well-sampled rows. A single feature builder, shared with evaluation, turns
these statistics into the per-edge matrix $Z$ whose rows hold the log visit
count $\log(1 + N_{s,a})$, the normalized mean reward $\hat r_{s,a}$, and the
empirical transition row $\hat P(\cdot \mid s,a)$ (uniform when the pair is
unvisited). Because the same builder runs at training and at test time, the
model's input distribution is identical in both.

\subsection{Architecture}

The model is a Graph Attention Network with hidden width $d$ that plans over the
MDP graph by iterating a weight-tied propagation layer. Each edge $(s,a)$ is
encoded from its scalar features, and the state and edge embeddings are
initialized by pooling outgoing edges:
\begin{align*}
  &e_{sa} = \phi_{\text{enc}}\!\big([\log(1+N_{sa}),\, \hat r_{sa}]\big) \\
  &h^{(0)}_{sa} = e_{sa} \\
  &h^{(0)}_{s} = \phi_{\text{init}}\!
  \Big( \tfrac{1}{|\mathcal{A}|}\textstyle\sum_{a} e_{sa} \Big),
\end{align*}
where $\phi_{\text{enc}}, \phi_{\text{init}}$ are MLPs and $e_{sa} \in
\mathbb{R}^d$. Note that $\hat P$ enters only as the attention bias below, not
through $\phi_{\text{enc}}$.

Each propagation step $\ell = 0, \dots, K-1$ is a multi-head successor attention
followed by two residual updates. With $H$ heads of width $d_h = d/H$, the query
of an edge $(s,a)$ combines its current state-action embedding with its edge
encoding, while keys and values come from the successor states $s'$:
\begin{align*}
  &q^{(i)}_{sa} = W^{(i)}_q [\,h^{(\ell)}_{sa};\, e_{sa}\,],\\
  &k^{(i)}_{s'} = W^{(i)}_k h^{(\ell)}_{s'} \\
  &v^{(i)}_{s'} = W^{(i)}_v h^{(\ell)}_{s'}
\end{align*}
Each $(s,a)$ attends over candidate successors $s'$ with logits that add the
scaled dot product to the log empirical transition probability, and aggregates
the corresponding values:
\begin{align}
  &\alpha^{(i)}_{sa}(s') = \frac{\langle q^{(i)}_{sa},\,
  k^{(i)}_{s'}\rangle}{\sqrt{d_h}} + \beta \log \hat P(s' \mid s,a)
  \label{eq:succ-attn}
\end{align}
and then
\begin{align*}
  &m_{sa} = W_o\, \operatorname{concat}\big(m^{(1)}_{sa}, \dots, 
  m^{(H)}_{sa}\big)\\
  &m^{(i)}_{sa} = \sum_{s'} \operatorname{softmax}_{s'}\!
  \big(\alpha^{(i)}_{sa}\big)\, v^{(i)}_{s'}
\end{align*}

The additive bias in \eqref{eq:succ-attn} is what ties the layer to planning,
and its strength $\beta$ acts as an attention regularizer that controls how
tightly the attention is pinned to the empirical dynamics (Table~\ref{tab:prior};
default $\beta = 1$). When the content logits $\langle q, k\rangle$ are constant
in $s'$, the weights reduce to $\operatorname{softmax}_{s'}(\beta \log \hat P)
\propto \hat P(\cdot \mid s,a)^{\beta}$: at $\beta = 1$ this is exactly $\hat
P(\cdot \mid s,a)$, so the message collapses to the Bellman expectation $m_{sa} =
\mathbb{E}_{s' \sim \hat P(\cdot \mid s,a)}[\,v_{s'}\,]$, while $\beta > 1$
sharpens and $\beta < 1$ flattens that distribution. The content term is a
learned, data-dependent reweighting of this backup, and the $\log \hat P = -
\infty$ entries suppress unreachable successors. The embeddings are then refreshed
with residual MLPs and LayerNorm,
\begin{align*}
  h^{(\ell+1)}_{sa} &= \operatorname{LN}\!\big(h^{(\ell)}_{sa} + \phi_{sa}([\,h^{(\ell)}_{sa};\, m_{sa};\, e_{sa}\,])\big), \\
  h^{(\ell+1)}_{s}  &= \operatorname{LN}\!\Big(h^{(\ell)}_{s} + \phi_{s}\big([\,h^{(\ell)}_{s};\, \tfrac{1}{|\mathcal{A}|}\textstyle\sum_{a} h^{(\ell+1)}_{sa}\,]\big)\Big).
\end{align*}
After $K$ steps a readout produces a scalar logit per edge, and the policy is its
masked softmax over the valid actions $\mathcal{A}(s)$:
\begin{align*}
  u_\theta(s,a) = \phi_{\text{head}}\!\big([\,h^{(K)}_{s};\, h^{(K)}_{sa};\, e_{sa}\,]\big),
  \qquad
  \pi_\theta(a \mid s) = \frac{\exp u_\theta(s,a)}{\sum_{a' \in \mathcal{A}(s)} \exp u_\theta(s,a')}.
\end{align*}
The head is read only through this softmax, so it outputs a policy rather than
action values. Because the propagation layer is weight-tied across the $K$
iterations, the network can be unrolled to a greater depth at evaluation than
during training.

\paragraph{Padding and masking.} MDPs in a batch vary in $|\mathcal{S}|$ and
$|\mathcal{A}|$, so each is padded to the fixed maxima $S_{\max} = 32$ and
$A_{\max} = 4$. A boolean state and action mask marks the valid rows: padded
edges are zeroed in the encoder and after every propagation update, padded
successors are dropped from the attention, and invalid actions receive a large
negative logit before the policy softmax. Padding therefore never contributes to
a valid row's representation, and the loss is averaged over valid states only.

\subsection{Training objective and optimization}

Each gradient 
step resamples a fresh batch of MDPs and their $Z$ matrices. For a
sampled MDP $M$, value iteration on the true dynamics gives $Q^*$, and the
per-MDP scale $\rho = \max_{s,a} |r(s,a)|$ defines the temperature-$\tau$
Boltzmann target policy
\begin{align*}
  \pi^*_\tau(a \mid s) = \operatorname{softmax}_{a \in \mathcal{A}(s)}\!\Big( \frac{Q^*(s,a)}{\rho\,\tau} \Big).
\end{align*}
The objective is the expected per-state cross-entropy between this target and the
model's prediction,
\begin{align}
  \mathcal{L}(\theta) = \mathbb{E}_{M,\,Z \sim \mathrm{explore}(M)}
  \Bigg[\, \frac{1}{|\mathcal{S}|} \sum_{s \in \mathcal{S}}
    \Big( -\!\!\sum_{a \in \mathcal{A}(s)} \pi^*_\tau(a \mid s)\, \log \pi_\theta(a \mid s;\, Z) \Big)
  \Bigg],
  \label{eq:loss}
\end{align}
averaged over valid states and over the batch. Since cross-entropy with a soft
target is minimized, for each fixed $Z$, at the conditional mean, the population
minimizer of \eqref{eq:loss} is the posterior mean policy $\pi_\theta(\cdot \mid
s;\, Z) = \mathbb{E}[\pi^*_\tau(\cdot \mid s) \mid Z]$, the Bayes-optimal estimate
of Section~\ref{sec:arch}. The KL between $\pi^*_\tau$ and $\pi_\theta$ is logged
as a diagnostic only. We optimize with AdamW under a linear warmup followed by
cosine decay with gradient-norm clipping. Default settings: hidden width $256$, $8$
attention heads, $K = 20$ propagation steps, dropout $0.05$, batch size $128$,
$\tau = 0.2$, learning rate $3 \times 10^{-4}$, weight decay $0.01$, and
gradient clip $0.5$.

\section{Evaluation protocols}
\label{app:eval}

Both evaluations share the same autoregressive rollout. The model is never shown
the true $(P, r)$. It starts from a uniform transition estimate and zero rewards,
and once per episode it re-plans a policy $\pi_t$ from the statistics $Z_t$
gathered so far. That policy is then held fixed for the episode while the
experience it generates updates the counts that form $Z_{t+1}$. Every episode
starts at $s_{\mathrm{start}} = 0$ and runs for $50$ steps. Exploration samples
actions from $\operatorname{softmax}(u_\theta / \tau)$ with temperature $\tau$;
the episodes we actually measure are run greedily 
($\tau \to 0$, strict $\arg\max$) so that the recorded number reflects 
exploitation rather than exploration. The two protocols differ only in how a 
policy is scored.

\subsection{Benchmarks}
The two benchmarks used are GridWorld~\citep{sutton2018reinforcement} and 
FrozenLake~\citep{brockman2016openai}. 

For GridWorld, the initial and final states are always respectively the top-left
and bottom-right corners of the square. We set $\texttt{step\_cost} = -1$ and
$\texttt{goal\_r} = 10.0$.

For FrozenLake, we use a $4 \times 4$ grid, with holes at index $5, 7, 11, 12$
(the grid is indexed row-major), $\texttt{slip} = 0.2$,
$\texttt{step\_cost} = 0$, $\texttt{hole\_r} = -1$ and $\texttt{goal\_r} = 1.0$.

\subsection{Learning curves (Figure~\ref{K_evals})}

The first protocol measures the greedy return as a function of the number of
episodes in context and of the evaluation depth $K$. We roll out $512$ episodes
per environment, built with $\tau = 0.3$ and read off the greedy return 
at every episode index that is a power of two, 
sweeping $K \in \{4, 8, 16, 20, 24, 38\}$ over the weight-tied
propagation layer. A greedy episode is scored by its return $R$: the sum of
rewards collected over the $50$ steps, with no discounting. FrozenLake is
stochastic, so a single rollout is noisy; we average $256$ greedy rollouts per
measured point and let only the first of them write to the statistics. The
deterministic GridWorlds use a single rollout.

Returns are normalized against two references estimated by the same rollout,
each averaged over $2000$ episodes: $R_{\mathrm{opt}}$ from the value-iteration
policy and $R_{\mathrm{rand}}$ from the uniform-random policy. The reported score
is $(R - R_{\mathrm{rand}}) / (R_{\mathrm{opt}} - R_{\mathrm{rand}})$, so a random
policy sits at $0$ and the optimal policy at $1$. Because both the numerator and
$R_{\mathrm{opt}}$ are sampled, a near-optimal policy can land slightly above $1$
on noise.

\subsection{Episodes to convergence (Table~\ref{tab:qlearning_ucb})}

The second protocol measures how many episodes the greedy policy needs before it
is optimal, and compares that count against UCB-VI and tabular Q-learning. The 
score here is exact rather than sampled. At each episode we take
the greedy policy $\pi_t = \arg\max_a u_\theta(s, a)$ and solve the linear system
$V^{\pi_t} = (I - \gamma P_{\pi_t})^{-1} r_{\pi_t}$ for its discounted value at 
the start state. With $V^*$ from value iteration and $V_{\mathrm{rand}}$ the 
exact value of the uniform-random policy, the normalized score is
$(V^{\pi_t}(s_{\mathrm{start}}) - V_{\mathrm{rand}}) /
(V^*(s_{\mathrm{start}}) - V_{\mathrm{rand}})$. The greedy policy counts as
converged once this score holds at or above the threshold for $8$ consecutive
episodes, so a single $\arg\max$ flip from an agent that is still exploring
resets the count. Because the episodes we score are greedy regardless, we raise the exploration
temperature to $\tau = 1.0$ when collecting data. 
The reported figure is the first episode of that window,
taken as the median over $12$ seeds. The two tabular baselines are scored the
same way, episode by episode: UCB-VI plans on its empirical model plus a
Hoeffding bonus with $c = 1$, and Q-learning is $\varepsilon$-greedy, with 
optimistic initialization, where $\varepsilon = 0.1$.

The exact discounted value is a stricter target than the sampled return of the
figure. On the deterministic GridWorlds the model recovers the optimal policy
outright, so both protocols agree and the score is exactly $1$. On FrozenLake 
the in-context policy settles just under the optimum in exact value. The table 
is therefore sensitive to the threshold: at $0.95$ it converges within tens of
episodes, while a threshold near $1$ would never register, even though the
sampled curve in the figure already looks saturated.

\subsection{Offline policy recovery (Figure~\ref{fig:offline})}

The third protocol isolates the model as an offline estimator: given a fixed
dataset of transitions, how good a policy can it recover, and how does that
compare against a standard offline-RL baseline on the identical data. We fix the
behavior policy to uniform random, reset to $s_{\mathrm{start}} = 0$ whenever an
absorbing state is reached or after $50$ steps, and collect a single stream of
transitions per seed. At every dataset size that is a power of two (from $8$ to
$2048$ transitions) we freeze the accumulated statistics, the counts
$N(s,a,s')$, the summed rewards, and the observed reward scale, and feed the
identical $Z$ to both estimators. Each produces a policy, scored by the exact
discounted normalized start value of the previous protocol. The curves are the
mean over $8$ seeds with one standard deviation shaded.

The baseline is pessimistic value iteration
\citep[VI-LCB,][]{rashidinejad2021bridging}, the offline counterpart of
UCB-VI. It plans on the maximum-likelihood model built from the counts,
subtracting a Hoeffding penalty $c\,\sqrt{\log(|\mathcal{S}||\mathcal{A}|/\delta)
/ N(s,a)}$ from each reward so that under-visited pairs are discouraged; pairs
never seen in the data are pinned to the pessimistic value floor
$-r_{\mathrm{scale}} / (1 - \gamma)$, and absorbing states are given continuation
value $0$, since their termination is recorded in the dataset. The in-context
model sees the same $Z$ and acts greedily.

We hand-picked $c$ to maximize performance on FrozenLake. A sweep over
$c \in \{0, 0.1, 0.3, 0.5, 1, 2\}$ left the deterministic GridWorlds at the
optimum for every value, but changed FrozenLake sharply: there the penalty
over-suppresses the rare observed goal transitions, so a large $c$ drives the
recovered policy below random. We report $c = 0.1$, the value that maximizes the
FrozenLake offline score. FrozenLake is the hard case for any data-only method
here: under uniform exploration from a fixed start the goal sits behind the
holes and is observed only a handful of times in thousands of transitions, so
VI-LCB is coverage-limited while the model can fall back on its prior.

\end{document}